\documentclass[journal]{IEEEtran}

\usepackage[utf8]{inputenc}
\usepackage[english]{babel}
\usepackage{amsthm} 

\usepackage{algorithm}
\usepackage{algpseudocode}

\usepackage{amssymb}  
\usepackage{amsmath}   
\usepackage{graphicx}  
\usepackage{subcaption}  
\usepackage{multirow}  
\usepackage{xcolor}  

\usepackage{url}
\usepackage{booktabs,tabularx}

\ifCLASSINFOpdf
\else
\fi
\usepackage{eso-pic} 

\begin{document}

\AddToShipoutPictureBG*{%
  \AtPageUpperLeft{%
    \setlength\unitlength{1in}%
    \hspace*{\dimexpr0.5\paperwidth\relax}
    \makebox(0,-0.75)[c]{\normalsize Published in IEEE Access, Volume 13, Pages 190582-190589, 2025, DOI: 10.1109/ACCESS.2025.3628158}
    }}

%
\title{Self-Supervised Learning \\Using Nonlinear  Dependence}
%
%
%

\author{M.Hadi Sepanj$^1$,
        Benyamin Ghojogh$^2$,
        Paul Fieguth$^1$ \\ 
        \hfil \break 
        $^1$Vision and Image Processing Group, Systems Design Engineering, University of Waterloo, Ontario, Canada \\
        $^2$Artificial Intelligence Scientist, Waterloo, Ontario, Canada \\
        Emails: \{mhsepanj, bghojogh, paul.fieguth\}@uwaterloo.ca
        }

%
%

\markboth{}%
{Shell \MakeLowercase{\textit{et al.}}: Bare Demo of IEEEtran.cls for IEEE Journals}
%



\maketitle

\begin{abstract}
Self-supervised learning has gained significant attention in contemporary applications, particularly due to the scarcity of labeled data. While existing SSL methodologies primarily address feature variance and linear correlations, they often neglect the intricate relations between samples and the nonlinear dependencies inherent in complex data--especially prevalent in high-dimensional visual data. In this paper, we introduce Correlation-Dependence Self-Supervised Learning (CDSSL), a novel framework that unifies and extends existing SSL paradigms by integrating both linear correlations and nonlinear dependencies, encapsulating sample-wise and feature-wise interactions. Our approach incorporates the Hilbert-Schmidt Independence Criterion (HSIC) to robustly capture nonlinear dependencies within a Reproducing Kernel Hilbert Space, enriching representation learning. Experimental evaluations on diverse benchmarks demonstrate the efficacy of CDSSL in improving representation quality.
\end{abstract}

\begin{IEEEkeywords}
Machine Learning, Deep Learning, Nonlinear Dependence, Self-Supervised Learning, Hilbert–Schmidt Independence Criterion, Reproducing Kernel Hilbert Space, Representation Learning.
\end{IEEEkeywords}

%
\IEEEpeerreviewmaketitle

\section{Introduction}
\label{sec:introduction}

Self-supervised learning (SSL) has revolutionized machine learning by enabling models to learn meaningful representations without requiring labeled data \cite{gui2024survey, hadi2025sinsim}. SSL methods achieve this by defining auxiliary tasks that leverage the structure within the data itself.
These representations are crucial for downstream tasks and must balance robustness, expressiveness, and transferability \cite{liu2021self}. A key challenge in SSL is minimizing redundancy and dependencies in the learned features while retaining relevant information \cite{shwartz2024compress}.

Well-known works, such as BYOL \cite{grill2020bootstrap}, SimCLR \cite{chen2020simple}, and SwAV \cite{caron2020unsupervised}, have proposed frameworks for learning representations by comparing multiple views of the same data. SimSiam \cite{chen2021exploring}, Barlow Twins \cite{zbontar2021barlow}, W-MSE \cite{ermolov2021whitening}, and VICReg \cite{bardes2022vicreg} focus on reducing feature redundancy through cross-correlation minimization and covariance regularization. While these methods have achieved significant success, they are often limited to {\em linear} dependencies, whereas {\em nonlinear} dependence reduction has recently gained attention as a way to improve representation quality, as explored in approaches like SSL in a Reproducing Kernel Hilbert Space (RKHS) \cite{li2021self, wu2025pseudo, ni2024graph}. In computer vision, the high intra-class variance and complex inter-class boundaries, often non-linearly separable in pixel space, make capturing nonlinear dependencies essential for learning robust representations \cite{chen2020simple,chen2021exploring}.


\begin{figure*}[!h]
\centering
\includegraphics[width=\textwidth]{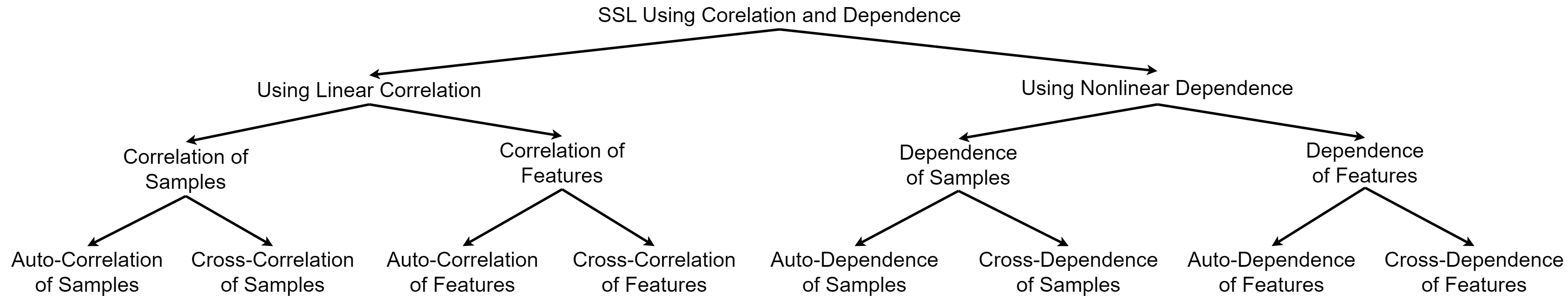}
\caption{Overview of the CDSSL framework. This figure highlights the novel categorization of dependencies into linear correlation and nonlinear dependence, further divided into sample-wise and feature-wise interactions. By addressing auto- and cross-dependence at both levels, CDSSL unifies existing SSL methods while introducing robust measures to enhance the quality of learned representations. This comprehensive framework bridges gaps in current methods, ensuring diversity and disentanglement in feature learning.}
\label{figure_SSL_categories}
\end{figure*}

In this work, we propose Correlation-Dependence SSL (CDSSL), a general framework for self-supervised learning that unifies and extends existing methods. CDSSL introduces eight loss terms that comprehensively address various aspects of correlation and dependence in SSL. As illustrated in Fig. \ref{figure_SSL_categories}, our framework systematically categorizes dependencies into \textit{linear} correlation and \textit{nonlinear} dependence, with further subdivisions into sample-wise and feature-wise corelation/dependence. These categories are then split into auto-correlation/dependence, which focuses on the representation of a batch of data, and cross-correlation/dependence, which examines the relationships between representations from two augmented views. By addressing dependencies at multiple levels, shown in Fig. \ref{figure_SSL_categories}, CDSSL ensures the learned features exhibit both diversity and disentanglement. Central to the nonlinear dependence aspect of CDSSL is the use of the Hilbert-Schmidt Independence Criterion (HSIC) to measure nonlinear dependencies, through RKHS, complementing traditional methods like InfoNCE \cite{oord2018representation}, which focus on similarity-based objectives.

Our contributions are fourfold: (1) we introduce a unified framework that categorizes dependencies in self-supervised learning (SSL) along linear/nonlinear, sample/feature, and auto/cross dimensions (Fig.~\ref{figure_SSL_categories}); (2) we incorporate HSIC to capture nonlinear dependencies via RKHS; (3) we show that CDSSL improves representation quality across multiple benchmarks; and (4) we demonstrate that several existing SSL methods—including VICReg, Barlow Twins, SimCLR, InfoNCE, and SSL-HSIC—can be seen as special cases of CDSSL.

\section{Background}


\subsection{Correlation}

Consider two random variables $X$ and $Y$ with samples $\{\mathbf{x} \in \mathbb{R}^d\}_{i=1}^n$ and $\{\mathbf{y} \in \mathbb{R}^d\}_{i=1}^n$, respectively. The correlation measures the linear dependence of the two random variables, and the widely-used Pearson correlation coefficient:
\begin{align}\label{equation_pearson_correlation}
\textbf{Corr}(X, Y) = \frac{\textbf{Cov}(X, Y)}{\sqrt{\textbf{Var}(X)} \sqrt{\textbf{Var}(Y)}},
\end{align}
is a normalized form of the covariance. If $\mathbf{x}_{i,[j]}$ denotes the $j$-th coordinate of $\mathbf{x}_i$, then in principle we have both the sample correlation between samples,
\begin{align}\label{equation_pearson_correlation_samples}
& \textbf{Corr}(X_{i}, Y_{j}) := \frac{\sum_{k=1}^d \mathbf{x}_{i,[k]} \mathbf{y}_{j,[k]}}{\sqrt{\sum_{k=1}^d (\mathbf{x}_{i,[k]})^2} \sqrt{\sum_{k=1}^d (\mathbf{y}_{j,[k]})^2}}, 
\end{align}
and the sample correlation between features,
\begin{align}\label{equation_pearson_correlation_features}
& \textbf{Corr}(X_{[i]}, Y_{[j]}) := \frac{\sum_{k=1}^n \mathbf{x}_{k,[i]} \mathbf{y}_{k,[j]}}{\sqrt{\sum_{k=1}^n (\mathbf{x}_{k,[i]})^2} \sqrt{\sum_{k=1}^n (\mathbf{y}_{k,[j]})^2}} .
\end{align}
These correlations are represented as $(n \times n)$ and $(d \times d)$ matrices, respectively.

\subsection{Hilbert-Schmidt Independence Criterion (HSIC)}


The correlation is a fairly trivial measure of the inter-dependence of two random variables,  since it only captures {\it linear} dependence, whereas two variables are independent if and only if any bounded continuous ({\it nonlinear}) functions of them are uncorrelated. Therefore, samples from two random variables $\{\mathbf{x}\}_{i=1}^n$ and $\{\mathbf{y}\}_{i=1}^n$ can be mapped to different RKHSs using pulling functions \cite{paulsen2016introduction} $\phi(\mathbf{x})$ and $\phi(\mathbf{y})$. The correlation between $\phi(\mathbf{x})$ and $\phi(\mathbf{y})$ is computed in the Hilbert space \cite{alpay1984hilbert} to estimate their dependence in the input space \cite{ghojogh2023background}.
This operation is performed by the Hilbert-Schmidt Independence Criterion (HSIC), which measures the dependence of two random variables by computing the Hilbert-Schmidt norm of their cross-covariance \cite{gretton2005measuring}.

An empirical estimation of the HSIC is introduced in \cite{gretton2005measuring} since exact computation is intractable for finite datasets. The practical form widely used in machine learning is:
\begin{align}\label{equation_HSIC}
{\text{\textbf{HSIC}}}(X,Y) \approx \frac{1}{(n-1)^2}\, \textbf{tr}(\mathbf{K}_x\mathbf{H}\mathbf{K}_y\mathbf{H}),
\end{align}
where $\textbf{tr}(\cdot)$ is the trace, $\mathbf{H} := \mathbf{I} - (1/n) \mathbf{1}_n\mathbf{1}_n^\top$ is the centering matrix, and $\mathbf{K}_x := \mathbf{\phi}(\mathbf{x})^\top \mathbf{\phi}(\mathbf{x})$ and $\mathbf{K}_y := \mathbf{\phi}(\mathbf{y})^\top \mathbf{\phi}(\mathbf{y})$ are the kernels for $\mathbf{x}$ and $\mathbf{y}$, respectively. 
The normalization term $1/(n-1)^2$ ensures consistency across datasets and can be omitted for comparative analyses.

${\text{\textbf{HSIC}}} = 0$ indicates that $\mathbf{x}$ and $\mathbf{y}$ are independent, while ${\text{\textbf{HSIC}}} > 0$ suggests a dependence between $\mathbf{x}$ and $\mathbf{y}$. A larger ${\text{\textbf{HSIC}}}$ value corresponds to a stronger nonlinear dependence, making it a reliable measure of such dependencies \cite{gretton2005measuring}. It can be shown that two random variables are independent if and only if their HSIC equals zero when using a universal kernel.


The HSIC of a random variable with itself, i.e., ${\text{\textbf{HSIC}}}(X,X)$, reduces to the covariance of the pulled data into one RKHS \cite{li2021self}. In this context, self-HSIC is not used to measure dependence between two random variables but rather to quantify the total variance, variability, or spread of a single random variable in the feature space (RKHS). This highlights another important property of HSIC: its ability to reflect the richness or complexity of data distributions. This perspective differs from the dependence-focused interpretation of HSIC and emphasizes its versatility in analyzing data characteristics beyond dependency.

\section{Network and Data Settings}\label{section_noncorresponding_samples}


Given a $d$-dimensional dataset $\{\mathbf{x}_\ell\}_{\ell=1}^n \subset \mathbb{R}^d$, the goal of SSL is to learn an embedding space where similar instances are close and dissimilar ones are far apart—but without using labels. This is achieved through data augmentation.

A typical SSL pipeline includes an encoder followed by an expander. The encoder $f_\theta$ maps inputs to a $q$-dimensional latent space, and the expander $h_\psi$ projects them to a higher-dimensional space $\mathbb{R}^p$ (with $p \gg q$):
\begin{align}
& f_\theta: \mathbb{R}^d \rightarrow \mathbb{R}^q, \quad \mathbf{y} := f_\theta(\mathbf{x}), \\
& h_\psi: \mathbb{R}^q \rightarrow \mathbb{R}^p, \quad \mathbf{z} := h_\psi(\mathbf{y}).
\end{align}
%
%
where $\theta$ and $\psi$ are learnable parameters. 
In SSL, each batch contains two sets of instances, $\{\mathbf{x}_i\}_{i=1}^b$ and $\{\mathbf{x}'_i\}_{i=1}^b$, where each pair $(\mathbf{x}_i, \mathbf{x}'_i)$ consists of different augmentations of the same original sample. Formally:
\begin{align}
\mathbf{x}_i := \mathcal{A}(\mathbf{x}_\ell), \quad \mathbf{x}'_i := \mathcal{A}(\mathbf{x}_\ell), \quad i \in \{1, \dots, b\},
\end{align}
with $\ell$ randomly chosen from $\{1, \dots, n\}$ and $\mathcal{A}(\cdot)$ denoting an augmentation such as rotation, cropping, scaling, or translation. As a result, each pair $(\mathbf{x}_i, \mathbf{x}'_i)$ shares the same underlying pattern, producing corresponding embeddings $(\mathbf{z}_i, \mathbf{z}'_i)$ without using class labels.

When the number of classes $c$ is much larger than the batch size $b$, the chance of sampling multiple instances from the same class is low. The probability of having at least $k$ samples from the same class in a batch is:
\begin{align}
\mathbb{P} = \sum_{i=k}^b {b \choose i} \left(\frac{1}{c}\right)^i \left(1-\frac{1}{c}\right)^{b-i},
\end{align}
which becomes negligible for large-$c$ datasets. Hence, non-corresponding pairs $(\mathbf{x}_i, \mathbf{x}_j)$ are likely from different classes, leading to distinct embeddings $(\mathbf{z}_i, \mathbf{z}_j)$.
This assumption underlies many contrastive SSL methods~\cite{oord2018representation,li2024enhancing} and few-shot learning approaches~\cite{khodadadeh2019unsupervised}, though it may break down when the number of classes is small.

\section{Linear Correlation Loss Terms}\label{section_linear_correlation_loss_terms}

In this section, we introduce the four linear correlation loss terms of CDSSL—auto‐ and cross‐correlation at both sample‐ and feature‐levels—which together serve to decorrelate irrelevant variation while aligning corresponding views across augmentations.
Let \(\widehat{\mathbf{z}}_{[k]}\) and \(\widehat{\mathbf{z}}'_{[k]}\) denote the \(k\)-th feature vectors, centered across the batch, computed as $\widehat{\mathbf{z}}_{[k]} := \mathbf{z}_{[k]} - (1/p) \sum_{\ell=1}^p \mathbf{z}_{[\ell]}$ and $\widehat{\mathbf{z}}'_{[k]} := \mathbf{z}'_{[k]} - (1/p) \sum_{\ell=1}^p \mathbf{z}'_{[\ell]}$.
Let \(\widetilde{\mathbf{z}}_k\) and \(\widetilde{\mathbf{z}}'_k\)  denote the embedding samples, centered across the batch, defined as $\widetilde{\mathbf{z}}_k := \mathbf{z}_k - (1/b) \sum_{\ell=1}^b \mathbf{z}_\ell$ and $\widetilde{\mathbf{z}}'_k := \mathbf{z}'_k - (1/b) \sum_{\ell=1}^b \mathbf{z}'_\ell$. 
The correlation between these vectors can be computed using Eqs. (\ref{equation_pearson_correlation_samples}) and (\ref{equation_pearson_correlation_features}).


\subsubsection{Auto-Correlation of Samples}
The correlation between the embeddings \(\mathbf{z}_i\) and \(\mathbf{z}_j\), \(i \neq j\), should be minimized because \(\mathbf{x}_i\) and \(\mathbf{x}_j\) have different patterns. Likewise, the correlation of \(\mathbf{z}'_i\) and \(\mathbf{z}'_j\), \(i \neq j\), should be minimized. The following \textit{auto-correlation loss between samples} should be minimized:
\begin{equation}
\label{equation_loss_auto_correlation_samples}
\begin{aligned}
\mathcal{L}_{\text{acs}} := \sum_{i=1}^b \sum_{j=1, j \neq i}^b \text{Corr}(\widetilde{\mathbf{z}}_i, \widetilde{\mathbf{z}}_j)^2 + \sum_{i=1}^b \sum_{j=1, j \neq i}^b \text{Corr}(\widetilde{\mathbf{z}}'_i, \widetilde{\mathbf{z}}'_j)^2,
\end{aligned} 
\end{equation}
where \(\widetilde{\mathbf{z}}_i\) and \(\widetilde{\mathbf{z}}'_i\) are the \(i\)-th centered embeddings.

\subsubsection{Cross-Correlation of Samples}
On the one hand, as the corresponding instances \(\mathbf{x}_i\) and \(\mathbf{x}'_i\) have similar patterns, the correlation of their output embeddings \(\mathbf{z}_i\) and \(\mathbf{z}'_i\) should be maximized. 
On the other hand, the cross-correlation between the non-corresponding embeddings \(\mathbf{z}_i\) and \(\mathbf{z}'_j\), for \(i \neq j\), should be minimized. Therefore, the following \textit{cross-correlation loss between samples} should be minimized:
\begin{equation}
\label{equation_loss_cross_correlation_samples}
\begin{aligned}
\mathcal{L}_{\text{ccs}} &:= \sum_{i=1}^b (1 - \text{Corr}(\widetilde{\mathbf{z}}_i, \widetilde{\mathbf{z}}'_i))^2  \\
&+ \eta_{\text{ccs}} \sum_{i=1}^b \sum_{j=1, j \neq i}^b \text{Corr}(\widetilde{\mathbf{z}}_i, \widetilde{\mathbf{z}}'_j)^2.
\end{aligned}
\end{equation}

\subsubsection{Auto-Correlation of Features}
The correlation between the embedding features \(\mathbf{z}_{[i]}\) and \(\mathbf{z}_{[j]}\), for \(i \neq j\), should be minimized to have low redundancy between the learned features. 
Likewise, the correlation of \(\mathbf{z}'_{[i]}\) and \(\mathbf{z}'_{[j]}\), for \(i \neq j\), should be minimized.
This decorrelates different features, within the batch, and it avoids collapse of the embeddings into zero vectors. The following \textit{auto-correlation loss between features} should be minimized:
\begin{equation}
  \label{equation_loss_auto_correlation_features} 
\begin{aligned}
\mathcal{L}_{\text{acf}} &:= \sum_{i=1}^p \sum_{j=1, j \neq i}^p \text{Corr}(\widehat{\mathbf{z}}_{[i]}, \widehat{\mathbf{z}}_{[j]})^2 \\
&+ \sum_{i=1}^p \sum_{j=1, j \neq i}^p \text{Corr}(\widehat{\mathbf{z}}'_{[i]}, \widehat{\mathbf{z}}'_{[j]})^2,
\end{aligned}
\end{equation}
where \(\widehat{\mathbf{z}}_{[i]} := [\widehat{\mathbf{z}}_{1,[i]}, \dots, \widehat{\mathbf{z}}_{b,[i]}]^\top\) is the vector of the \(i\)-th feature across the batch.

\subsubsection{Cross-Correlation of Features}
The corresponding embedding features \(\mathbf{z}_{[i]}\) and \(\mathbf{z}'_{[i]}\) should be similar to have the same embedding features for instances with similar patterns. Therefore, the correlation between them should be maximized. 
Additionally, to minimize redundancy between the learned features across augmentations, the correlation between the embedding features \(\mathbf{z}_{[i]}\) and \(\mathbf{z}'_{[j]}\), for \(i \neq j\), should be minimized. This decorrelates different features, across augmentations, and it avoids collapse of the embeddings into zero vectors. 
Overall, the following \textit{cross-correlation loss between features} should be minimized:
\begin{equation}
  \label{equation_loss_cross_correlation_features}
\begin{aligned}
\mathcal{L}_{\text{ccf}} &:= \sum_{i=1}^p (1 - \text{Corr}(\widehat{\mathbf{z}}_{[i]}, \widehat{\mathbf{z}}'_{[i]}))^2 \\
&+ \eta_{\text{ccf}} \sum_{i=1}^p \sum_{j=1, j \neq i}^p \text{Corr}(\widehat{\mathbf{z}}_{[i]}, \widehat{\mathbf{z}}'_{[j]})^2,
\end{aligned}
\end{equation}
where \(\widehat{\mathbf{z}}_{[i]}\) and \(\widehat{\mathbf{z}}'_{[j]}\) are the \(i\)-th and \(j\)-th feature vectors across the batch, respectively.

\section{Nonlinear Dependence Loss Terms}\label{section_nonlinear_dependence_loss_terms}

\subsubsection{Auto-Dependence of Samples}

The samples in a batch, $\{\mathbf{z}_i\}_{i=1}^b$ or $\{\mathbf{z}'_i\}_{i=1}^b$ should exhibit sufficient richness to represent the structure of the data. In other words, assuming that the instances in the batch belong to different classes, the variance of embeddings in the feature space should be maximized. 
This can be measured by self-HSIC (${\text{\textbf{HSIC}}}(X,X)$) which is also a measure of dependence. 
The following \textit{auto-dependence loss between samples} should be minimized:
\begin{equation}\label{equation_loss_auto_dependence_samples}
\begin{aligned}
\mathcal{L}_{\text{ads}} &:= - {\text{\textbf{HSIC}}}(\{\mathbf{z}_i\}_{i=1}^b, \{\mathbf{z}_i\}_{i=1}^b) - {\text{\textbf{HSIC}}}(\{\mathbf{z}'_i\}_{i=1}^b, \{\mathbf{z}'_i\}_{i=1}^b) \\
&= - \textbf{tr}(\mathbf{K}_{\mathbf{z}_i}\mathbf{H}_b\mathbf{K}_{\mathbf{z}_i}\mathbf{H}_b) - \textbf{tr}(\mathbf{K}_{\mathbf{z}'_i}\mathbf{H}_b\mathbf{K}_{\mathbf{z}'_i}\mathbf{H}_b),
\end{aligned}
\end{equation}
where $\mathbf{H}_b$ is the $(b \times b)$ centering matrix, and $\mathbf{K}_{\mathbf{z}_i} \in \mathbb{R}^{b \times b}$ and $\mathbf{K}_{\mathbf{z}'_i} \in \mathbb{R}^{b \times b}$ are the kernel matrices over $\{\mathbf{z}_i\}_{i=1}^b$ and $\{\mathbf{z}'_i\}_{i=1}^b$, respectively. 
Minimizing this loss function maximizes the self-HSIC of data and enhances the richness of data. 

\subsubsection{Cross-Dependence of Samples}

The dependence between the embeddings $\{\mathbf{z}_i\}_{i=1}^b$ and $\{\mathbf{z}'_i\}_{i=1}^b$ should be maximized, as $\{\mathbf{x}_i\}_{i=1}^b$ and $\{\mathbf{x}'_i\}_{i=1}^b$ share corresponding patterns due to augmentations. This ensures invariance to augmentation. The following \textit{cross-dependence loss between samples} should be minimized:
\begin{equation}\label{equation_loss_cross_dependence_samples}
\begin{aligned}
\mathcal{L}_{\text{cds}} &:= - {\text{\textbf{HSIC}}}(\{\mathbf{z}_i\}_{i=1}^b, \{\mathbf{z}'_i\}_{i=1}^b) = - \textbf{tr}(\mathbf{K}_{\mathbf{z}_i}\mathbf{H}_b\mathbf{K}_{\mathbf{z}'_i}\mathbf{H}_b).
\end{aligned}
\end{equation}
Minimizing this loss function maximizes the dependence of corresponding instances across augmentations. 


\subsubsection{Auto-Dependence of Features}

The embedding features $\{\mathbf{z}_{[i]}\}_{i=1}^p$ or $\{\mathbf{z}'_{[i]}\}_{i=1}^p$ should also exhibit sufficient richness to represent the structure of the data. To ensure diverse embedding features and prevent information collapse, the covariance of the embedding features should be maximized in the feature space, a property that can be quantified using self-HSIC. 
The following \textit{auto-dependence loss between features} should be minimized:
\begin{equation}\label{equation_loss_auto_dependence_features}
\begin{aligned}
&\mathcal{L}_{\text{adf}} := \\
&- {\text{\textbf{HSIC}}}(\{\mathbf{z}_{[i]}\}_{i=1}^p, \{\mathbf{z}_{[i]}\}_{i=1}^p) - {\text{\textbf{HSIC}}}(\{\mathbf{z}'_{[i]}\}_{i=1}^p, \{\mathbf{z}'_{[i]}\}_{i=1}^p) \\
&= - \textbf{tr}(\mathbf{K}_{\mathbf{z}_{[i]}}\mathbf{H}_p\mathbf{K}_{\mathbf{z}_{[i]}}\mathbf{H}_p) - \textbf{tr}(\mathbf{K}_{\mathbf{z}'_{[i]}}\mathbf{H}_p\mathbf{K}_{\mathbf{z}'_{[i]}}\mathbf{H}_p),
\end{aligned}
\end{equation}
where $\mathbf{H}_p$ is the $(p \times p)$ centering matrix and $\mathbf{K}_{\mathbf{z}_{[i]}} \in \mathbb{R}^{p \times p}$ and $\mathbf{K}_{\mathbf{z}'_{[i]}} \in \mathbb{R}^{p \times p}$ are the kernel matrices over $\{\mathbf{z}_{[i]}\}_{i=1}^p$ or $\{\mathbf{z}'_{[i]}\}_{i=1}^p$, respectively. 
Minimizing this loss function encourages the embedding features to capture diverse information, thereby preventing information collapse.

\subsubsection{Cross-Dependence of Features}

On the one hand, the dependence between the embedding features $\{\mathbf{z}_{[i]}\}_{i=1}^p$ and $\{\mathbf{z}'_{[i]}\}_{i=1}^p$ should be maximized to ensure invariant embedding features against augmentations. On the other hand, this dependence can be minimized if the features in one set are shuffled, promoting independence in such cases. The random permutations of features in one of them makes the structure of various features different from one another in a series of iterations. This reduces the redundancy of the learned embedding features. 
Overall, the following \textit{cross-dependence loss between features} should be minimized:
\begin{equation}\label{equation_loss_cross_dependence_features}
\begin{aligned}
&\mathcal{L}_{\text{cdf}} :=\\
&- {\text{\textbf{HSIC}}}(\{\mathbf{z}_{[i]}\}_{i=1}^p, \{\mathbf{z}'_{[i]}\}_{i=1}^p) + {\text{\textbf{HSIC}}}(\{\mathbf{z}_{[i]}\}_{i=1}^p, \{\mathbf{z}'_{\psi([i])}\}_{i=1}^p) \\
&= - \textbf{tr}(\mathbf{K}_{\mathbf{z}_{[i]}}\mathbf{H}_p\mathbf{K}_{\mathbf{z}_{[i]}}\mathbf{H}_p) + \textbf{tr}(\mathbf{K}_{\mathbf{z}'_{[i]}}\mathbf{H}_p\mathbf{K}_{\mathbf{z}'_{\psi([i])}}\mathbf{H}_p),
\end{aligned}
\end{equation}
where $\psi([i])$ is a random permutation of the features and $\mathbf{K}_{\mathbf{z}'_{\psi([i])}} \in \mathbb{R}^{p \times p}$ is the kernel matrix over $\{\mathbf{z}'_{\psi([i])}\}_{i=1}^p$. 
Minimizing this loss function makes the embedding features invariant to augmentations and also avoids redundancy in the features. 


It is noteworthy that we found out empirically that the embeddings, used in HSIC loss functions, are better to be normalized to have unit length. 

\section{Overall Loss Function}

The proposed loss functions are combined to make an overall loss function for CDSSL: 
\begin{equation}
    \label{equation_loss_total}
\begin{aligned}
\mathcal{L} &:= \underbrace{\lambda_{\text{acs}} \mathcal{L}_{\text{acs}} + \lambda_{\text{ccs}} \mathcal{L}_{\text{ccs}} + \lambda_{\text{acf}} \mathcal{L}_{\text{acf}} + \lambda_{\text{ccf}} \mathcal{L}_{\text{ccf}}}_{\text{linear correlation}} \\
&+ \underbrace{\lambda_{\text{ads}} \mathcal{L}_{\text{ads}} + \lambda_{\text{cds}} \mathcal{L}_{\text{cds}} + \lambda_{\text{adf}} \mathcal{L}_{\text{adf}} + \lambda_{\text{cdf}} \mathcal{L}_{\text{cdf}}}_{\text{nonlinear dependence}}, 
\end{aligned}
\end{equation}
where the $\lambda$ parameters are non-negative regularization hyperparameters. 
The first to fourth terms are linear correlation terms while the fifth to eighth terms are nonlinear dependence terms.

\begin{table*}[h]
    \centering
    \caption{Context and understanding of the CDSSL loss function terms.}
    \footnotesize
    \begin{tabular}{|c|p{16cm}|}
        \hline
        \textbf{Term} & \textbf{Description} \\ \hline
        $\mathcal{L}_{\text{acs}}$ & Decorrelates the embeddings of non-corresponding samples within each batch. \\ \hline
        $\mathcal{L}_{\text{ccs}}$ & Correlates corresponding samples across augmentations and decorrelates non-corresponding samples across augmentations. \\ \hline
        $\mathcal{L}_{\text{acf}}$ & Decorrelates non-corresponding embedding features within each batch, avoiding redundancy and information collapse. \\ \hline
        $\mathcal{L}_{\text{ccf}}$ & Correlates corresponding embedding features across augmentations and decorrelates non-corresponding features across augmentations, avoiding redundancy and information collapse. \\ \hline
        $\mathcal{L}_{\text{ads}}$ & Increases the variance of embeddings of non-corresponding samples within each batch in the feature space. \\ \hline
        $\mathcal{L}_{\text{cds}}$ & Increases the dependence of corresponding samples across augmentations. \\ \hline
        $\mathcal{L}_{\text{adf}}$ & Increases the covariance of embedding features within each batch, avoiding redundancy and information collapse through the feature space. \\ \hline
        $\mathcal{L}_{\text{cdf}}$ & Increases the dependence of corresponding embedding features across augmentations and decreases the dependence of non-corresponding features across augmentations, avoiding redundancy and information collapse. \\ 
        \hline
    \end{tabular}
    \label{table_cdssl_loss_analysis}
\end{table*}

\subsection{Analysis of the CDSSL Loss Function}

Analysis of the CDSSL loss function terms is provided in Table~\ref{table_cdssl_loss_analysis}.
The first term in the CDSSL loss function, $\mathcal{L}_{\text{acs}}$, decorrelates the embeddings of non-corresponding samples within each batch. The second term, $\mathcal{L}_{\text{ccs}}$, correlates the corresponding samples across augmentations and decorrelates the embeddings of non-corresponding samples across augmentations. The third term, $\mathcal{L}_{\text{acf}}$, decorrelates the non-corresponding embedding features within each batch, avoiding redundancy and information collapse. The fourth term, $\mathcal{L}_{\text{ccf}}$, correlates the corresponding embedding features across augmentations and decorrelates the non-corresponding embedding features across augmentations, avoiding redundancy and information collapse. 

The fifth term in the CDSSL loss function, $\mathcal{L}_{\text{ads}}$, increases the variance of the embeddings of non-corresponding samples within each batch in the feature space. The sixth term, $\mathcal{L}_{\text{cds}}$, increases the dependence of the corresponding samples across augmentations. The seventh term, $\mathcal{L}_{\text{adf}}$, increases the covariance of the embedding features within each batch which avoids redundancy and information collapse through the feature space. Finally, the last term, $\mathcal{L}_{\text{cdf}}$, increases the dependence of the corresponding embedding features across augmentations and decreases the dependence of the non-corresponding embedding features across augmentations, through the feature space, to avoid redundancy and information collapse. 

\subsection{Special Cases of CDSSL}\label{sup:special_cases}


Some existing SSL methods are special cases of the proposed CDSSL loss function. Specifically, methods based on variance, covariance, or correlation directly fit within our framework, whereas contrastive methods— which we conjecture—do so indirectly. In fact, many of the existing SSL methods can be mapped to one of the eight nodes in Fig. \ref{figure_SSL_categories}, making CDSSL a generalization of these approaches.

For instance, Barlow Twins \cite{zbontar2021barlow}, which uses cross-correlation of features, is a direct special case of CDSSL with only $\lambda_{\text{ccf}} = 1$ non-zero. Thus, CDSSL's performance is lower bounded by Barlow Twins. Similarly, VICReg \cite{bardes2022vicreg} includes variance and covariance terms that are implicitly equivalent to cross-correlation, albeit with different weights.

The SSL-HSIC loss includes both cross-dependence and auto-dependence among samples. Unlike CDSSL, however, it models only sample-level dependencies and omits any feature-level decorrelation. Hence, SSL-HSIC \cite{li2021self}, which also employs HSIC, can be regarded as a special case of the CDSSL loss function.


InfoNCE \cite{oord2018representation}, also used in SimCLR \cite{caron2020unsupervised}, is another special case of the CDSSL loss. As shown in \cite{li2021self}, InfoNCE can be viewed as a special case of SSL-HSIC with a cosine kernel. Since SSL-HSIC is itself a special case of CDSSL with only cross- and auto-dependence of samples, InfoNCE also fits within CDSSL under the same terms and kernel.

By combining multiple non-zero $\lambda$ values, CDSSL integrates the strengths of various SSL methods—such as Barlow Twins, VICReg, InfoNCE, SimCLR, and SSL-HSIC—while also introducing novel components. This synergy enhances the quality of the learned embeddings, unifying and extending existing approaches.

\section{Experimental Results}

In this section, we delve into the evaluation and performance analysis of the proposed approach.


\subsection{Hyperparameter Sensitivity Analysis}\label{sup:hyperparameter_sensitivity}

\begin{figure}[!t]
  \centering
  \includegraphics[width=.45\linewidth]{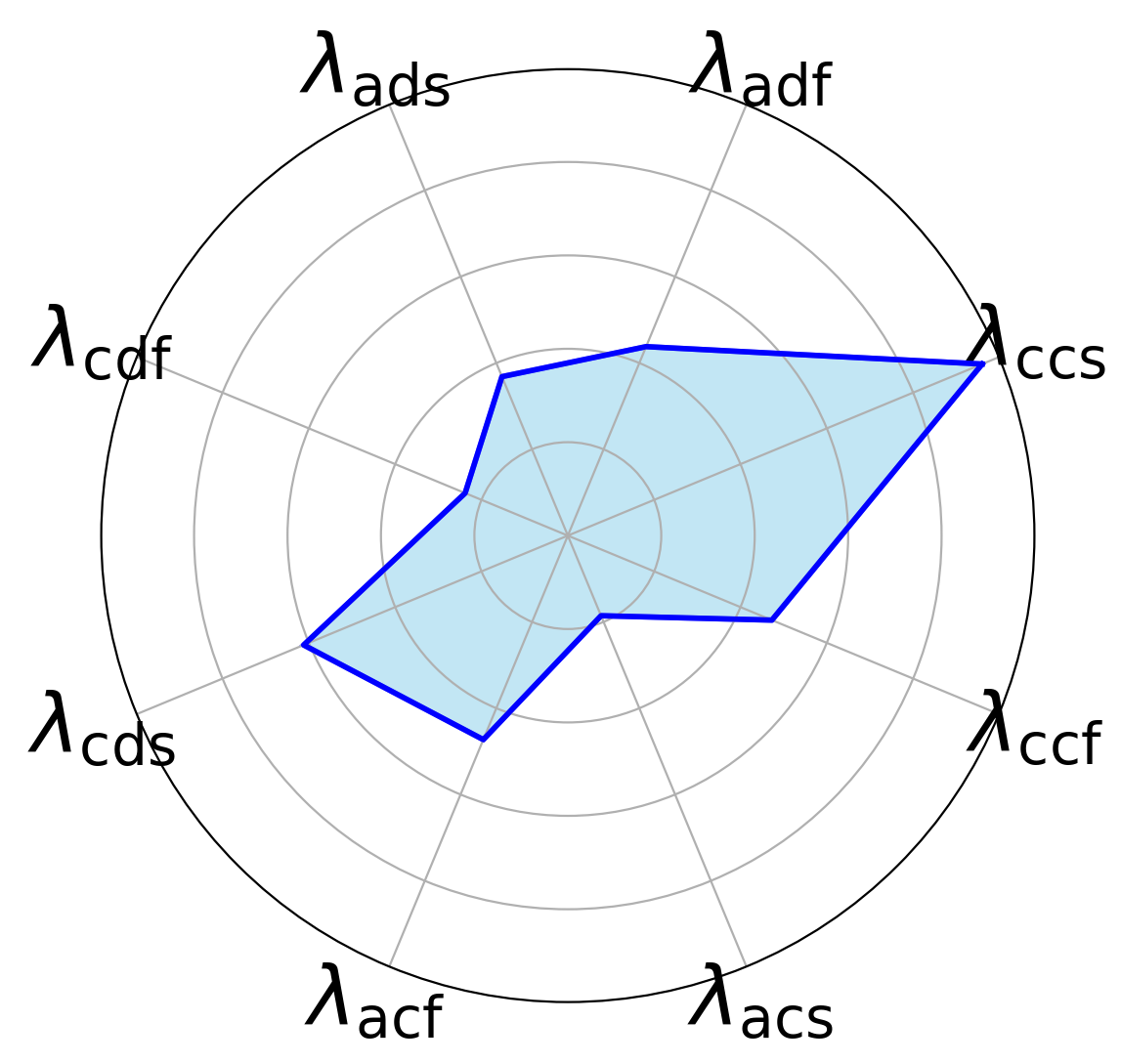}
  \includegraphics[width=.45\linewidth]{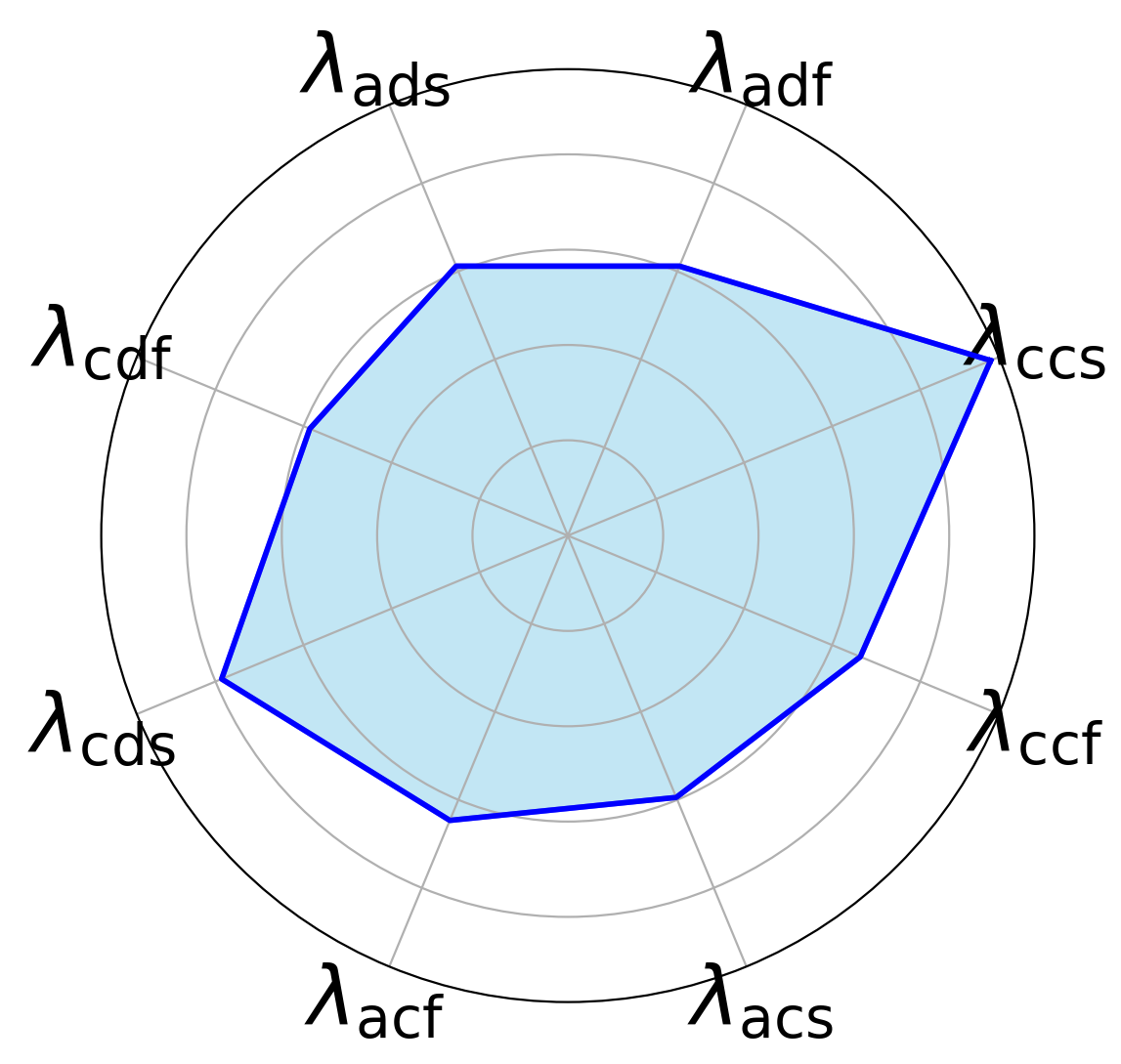}
  \caption{Optimal relative values of the regularization hyperparameters for the CDSSL loss function, obtained through grid search, on (a) MNIST with 10 classes (left figure) and (b) CIFAR-100 with 100 classes (right figure). The results highlight the varying importance of different loss terms based on the dataset's class complexity. Cross-correlation and cross-dependence of samples are more prominent in datasets with a larger number of classes.}
  \label{fig_ablation}
\end{figure}

To understand the sensitivity of CDSSL performance to hyperparameter choices, we conducted an analysis across datasets with few and many classes. 
The proposed CDSSL framework includes several regularization hyperparameters (see Eq.~(\ref{equation_loss_total})) which need to be configured. Setting these parameters is straightforward due to an intuitive understanding of each loss term (see Table~\ref{table_cdssl_loss_analysis}). In Fig.~\ref{fig_ablation}, we present results from a grid-based hyperparameter optimization conducted on two illustrative datasets representing distinct scenarios: one with few classes (MNIST) and one with many classes (CIFAR-100). Before the optimization, hyperparameters were standardized for a fair comparison between loss terms.

According to Fig.~\ref{fig_ablation}, cross-correlation and cross-dependence between samples are the most impactful terms, consistent with the objective of aligning embeddings of augmented samples. Notably, auto-correlation and auto-dependence of samples have reduced significance in scenarios with fewer classes (e.g., MNIST), as the assumption regarding distinctiveness of non-corresponding samples (see Section~\ref{section_noncorresponding_samples}) is less valid in such cases. Other loss terms demonstrate moderate and consistent influence across scenarios.

This analysis serves as practical guidance for setting hyperparameters when applying CDSSL to new datasets. One can assess which scenario—few or many classes—is closer and select hyperparameter values accordingly.

\begin{table*}[!t]
    \centering
    \small
    \caption{Performance comparison of self-supervised learning  on four benchmark datasets: MNIST, CIFAR-10, STL10, and CIFAR-100. Metrics are reported as top-1 and top-5 accuracy (\%). CDSSL outperforms VICReg and Barlow Twins across most of the datasets, demonstrating its effectiveness in learning meaningful and transferable representations.}
    \label{table_linear_classification_simpler}
    \footnotesize
    \begin{tabular}{l|*{4}{cc}}
        \toprule
        & \multicolumn{2}{c}{\textbf{MNIST}} 
        & \multicolumn{2}{c}{\textbf{CIFAR-10}} 
        & \multicolumn{2}{c}{\textbf{STL10}}
        & \multicolumn{2}{c}{\textbf{CIFAR-100}} 
        \\
        \cmidrule(lr){2-3} \cmidrule(lr){4-5} \cmidrule(lr){6-7} \cmidrule(lr){8-9}
        \textbf{Method} 
        & \textbf{Top-1} & \textbf{Top-5} 
        & \textbf{Top-1} & \textbf{Top-5} 
        & \textbf{Top-1} & \textbf{Top-5}
        & \textbf{Top-1} & \textbf{Top-5}
        \\
        \midrule
        \textbf{VICReg} 
        & 93.2 & 99.6 
        & 72.4 & 98.0 
        & \textbf{52.4} & \textbf{94.1} 
        & 45.5 & 74.8 \\
        \textbf{Barlow Twins}
        & 92.8 & 99.7 
        & 72.1 & 97.9
        & 51.3 & 93.1 
        & 46.1 & 75.1 \\
        \textbf{SimCLR}
        & \textbf{93.9} & 99.7
        & 68.6 & 95.9
        & 50.3  & 93.6
        & 41.9 & 68.9 \\
        \textbf{SSL-HSIC}
        & 92.5 & 99.6
        & 71.7 & 97.3
        & 51.9  & 93.3
        & 44.6 & 74.1 \\
        \textbf{CDSSL} 
        & 93.5 & \textbf{99.7}
        & \textbf{74.6} & \textbf{98.1} 
        & 51.5 & 93.2
        & \textbf{47.0} & \textbf{76.6}\\
        \bottomrule
    \end{tabular}
\end{table*}

\begin{table*}[!t]\small
    \centering
    \caption{Used hyperparameters in the experiments.}
    \label{tab_hyperparameters}
    \begin{tabular}{l|lllllllll}
        \toprule
        & $\lambda_{\text{acs}}$ & $\lambda_{\text{ccs}}$ & $\lambda_{\text{acf}}$ & $\lambda_{\text{ccf}}$ & 
        $\lambda_{\text{ads}}$ &
        $\lambda_{\text{cds}}$ & $\lambda_{\text{adf}}$ & $\lambda_{\text{cdf}}$ & \\
        \midrule
        MNIST & 0.00001 & 0.01 & 0.0001 & 0.001 & 1 & 110 & 0.1 & 1 \\
        CIFAR-10 & 0.0001 & 0.1 & 0.001 & 0.1 & 0.1 & 150 & 0.1 & 5  \\
        STL10 & 0.0001 & 0.1 & 0.001 & 0.1 & 1 & 200 & 0.1 & 6 \\
        CIFAR-100 & 0.0002 & 0.01 & 0.01 & 0.2 & 1.2 & 160 & 0.2 & 6  \\
        \bottomrule
    \end{tabular}
\end{table*}


\subsection{Evaluation by Linear Classification}

\subsubsection{Evaluation and Comparison}

All experiments in the paper have used the RBF kernel function, which is a universal kernel and thus is valid to be used in HSIC.
The proposed algorithm was trained and tested on well-known benchmarks: MNIST \cite{lecun1998gradient}, CIFAR-10, CIFAR-100 \cite{krizhevsky2009learning}, and STL10 \cite{coates2011analysis}. 

We tested on multiple datasets with different distributions and domains. Having different datasets with different distributions showcases the statistical significance of our performance. Moreover, among the datasets we use are cases with few and many classes.
We trained and tested VICReg \cite{bardes2022vicreg}, Barlow Twins \cite{zbontar2021barlow}, SSL-HSIC \cite{li2021self} and SimCLR \cite{chen2020simple} as methods to compare to, all implemented with consistent settings for fair comparison.  

After training the SSL method, the expander was dropped and a linear logistic regression was trained and tested on the learned latent embeddings. The top-1 and top-5 accuracies are reported in Table \ref{table_linear_classification_simpler}. The performance of CDSSL is better than VICReg, Barlow Twins, SSL-HSIC and SimCLR in most of the cases.

\subsubsection{Detailed Settings of Experiments}\label{section_settings}

For reproducibility of the results, the detailed settings of the experiments are reported in the following.
We have conducted experiments on different datasets which Table \ref{tab_hyperparameters} reports the corresponding $\lambda$ for each. 
Also, $\eta_{\text{ccs}}=\eta_{\text{ccf}}=0.05$ was used for all experiments.
Note that the ablation study in Fig. \ref{fig_ablation} reports $\lambda$ values after unifying the scales of loss terms for fair comparison and analysis of the hyperparameters. However, Table \ref{tab_hyperparameters} reports the actual values of hyperparameters for the datasets. 
In all of the experiments, we have employed a batch size of $512$ and trained the model for 500 epoch (except for MNIST which the number of epochs are 100). The linear classifier (logistic regression) is trained for 20 epochs. The selected backbone is a simple convolutional neural network for MNIST and ResNet-18 \cite{he2016deep} for the rest of the datasets. Dimensionality of the latent space is set to $512$ and the expander network has a dimension of $1024$ for CDSSL, VICReg, and Barlow Twins, and a dimension of $128$ for SimCLR. The Adam optimizer was used for optimization with a learning rate of $0.0002$ and a weight decay of $0.000001$.

\subsubsection{Visualization of the Learned Embeddings}

\begin{figure*}[h]
\centering
\includegraphics[width=\textwidth]{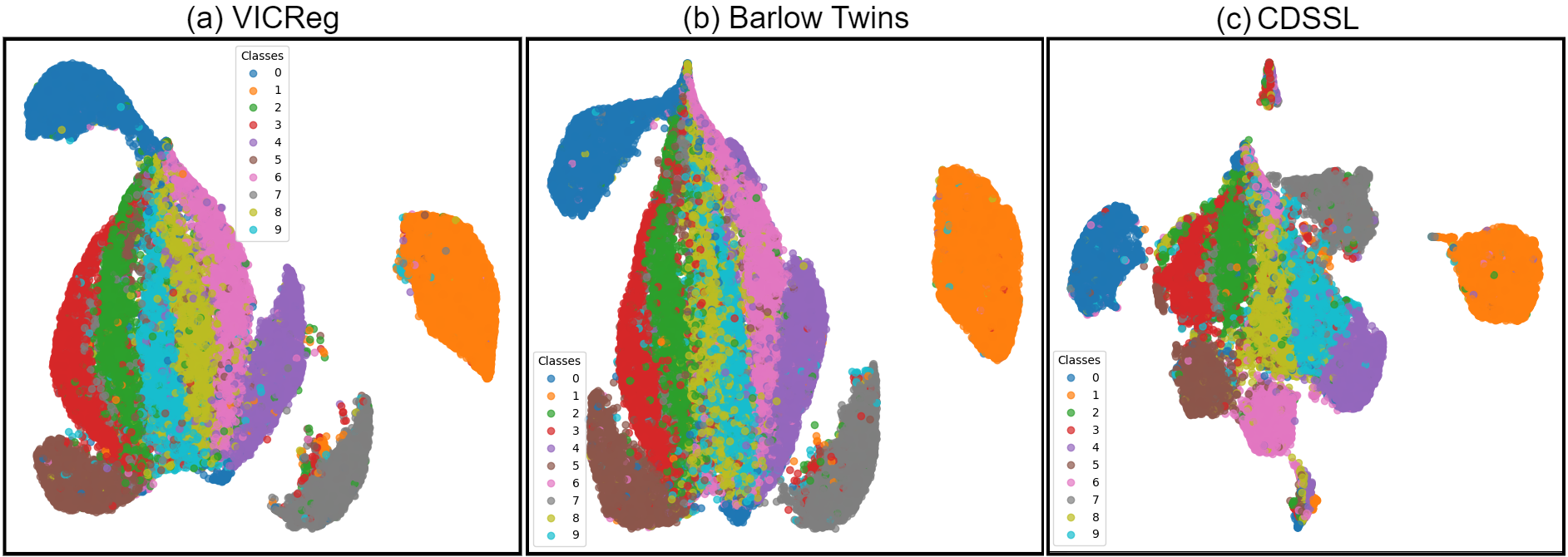}
\caption{Four UMAP visualizations of the learned embeddings for the MNIST dataset. CDSSL in (c) shows better class separation and more isotropic distributions compared to VICReg and Barlow Twins, highlighting its ability to capture discriminative representations.}
\label{fig_comparison_umap}
\end{figure*}

The UMAP visualizations \cite{mcinnes2018umap} of the learned latent embeddings by CDSSL, VICReg, and Barlow Twins are illustrated in Fig. \ref{fig_comparison_umap} for the MNIST dataset. The visualizations clearly show that the classes have been better separated and more isotropically distributed by CDSSL than VICReg and Barlow Twins. The reason for this behavior is explained in the following. 

CDSSL includes loss terms such as auto-correlation of samples and auto-dependence of samples that explicitly minimize the correlation/dependence among non-corresponding samples. This makes the classes separated in the embedding space while the features are also non-redundant because of the other loss terms. More inter-class variances results in better uniform propagation of the classes in the embedding space so their UMAP embedding shows isometric classes. This is while there is no explicit loss term in VICReg and Barlow Twins for repelling different (non-corresponding) samples from each other. Both VICReg and Barlow Twins work on the \textit{features} to be non-redundant (uncorrelated) and non-collapsing but they do not explicitly deal with the \textit{samples}. Hence, the inter-class variances are not as much as CDSSL.


\subsection{Experiments on Imagenet100}

We also evaluated on ImageNet-100 \cite{deng2009imagenet}; the results show that our method is comparable to other recent approaches—outperforming all except DINO \cite{caron2021emerging} by a slight margin.
As shown in Table~\ref{tab:imagenet100-ssl}, CDSSL achieves a kNN Top-1 accuracy of 51.1\% on ImageNet-100 with a ResNet-18 backbone \cite{he2016deep}, markedly outperforming Barlow Twins (46.5\%), BYOL (43.9\%), NNCLR (45.3\%) and SimCLR (46.9\%), and trailing only DINO (51.8\%) by a narrow 0.7 point margin. This strong kNN performance underlines the value of integrating nonlinear dependence via HSIC into our framework. 

\subsection{Evaluation by Nonlinear Classification}

The proposed CDSSL was compared with VICReg, Barlow Twins, and SimCLR by nonlinear classification on four datasets in Table \ref{table_nonlinear_classification}. The nonlinear classifier is a multilayer Perceptron with three layers. The resulting classifier, based on the learned latent embeddings, is strikingly superior for CDSSL, outperforming all of the baseline methods, supporting the generalization ability of CDSSL. 

\begin{table}[!t]
\centering
\caption{Comparison of SSL methods on ImageNet-100 with ResNet-18 backbone. All models were trained for 200 epochs with batch size 256.}
\label{tab:imagenet100-ssl}
\begin{tabular}{lcccccc}
\toprule
\textbf{Model} & \textbf{Backbone} & \textbf{Batch Size} & \textbf{Epochs} & \textbf{Top-1 KNN} \\
\midrule
Barlow Twins & Res18 & 256 & 200 & 0.465   \\
BYOL         & Res18 & 256 & 200 & 0.439   \\
DINO         & Res18 & 256 & 200 & 0.518  \\
NNCLR        & Res18 & 256 & 200 & 0.453 \\
SimCLR       & Res18 & 256 & 200 & 0.469  \\
CDSSL        & Res18 & 256 & 200 & 0.511   \\
\bottomrule
\end{tabular}
\end{table}

\begin{table}[!t]
    \centering
    \caption{Downstream comparison of nonlinear classification performance, on top of frozen representations.  CDSSL provides a strikingly stronger performance, relatively to all compared methods, and for every dataset.}
    \label{table_nonlinear_classification}
    \footnotesize
    \begin{tabular}{l|cccc}
        \toprule
        \textbf{Dataset} & \textbf{VICReg}  & \textbf{Barlow Twins} & \textbf{SimCLR}& \textbf{CDSSL} \\
        \midrule
        MNIST  & 95.90 & 94.96  & 96.27 & \textbf{97.24} \\
        CIFAR-10  & 74.52 & 73.62 & 70.85 & \textbf{75.37}  \\
        STL10  & 53.93 & 52.21 & 52.60 & \textbf{54.47}  \\
        CIFAR-100  & 46.21 & 46.07  & 42.11 &  \textbf{47.08}    \\
        
        \bottomrule
    \end{tabular}
\end{table}

\subsection{Domain Adaptation}


We evaluated the cross-dataset generalization capabilities of the proposed CDSSL framework using the domain adaptation tasks shown in Table~\ref{table_domain_adaptation}. In these experiments, models were trained on one dataset and the expander network discarded. A linear classifier was trained on top of the learned representations without fine-tuning to assess the generalization to unseen datasets.
The used datasets in this experiment are MNIST, Digits Dataset \cite{scikitlearn2011pedregosa}, CIFAR-10, STL10, and CIFAR-100.
Again, the results indicate that CDSSL consistently outperforms VICReg, Barlow Twins, and SimCLR across all tasks. 


\begin{table*}[!t]
    \centering
    \caption{Domain adaptation performance comparison between VICReg and CDSSL. 
    A linear classifier was trained on top of the frozen representations to evaluate cross-dataset generalization.}
    \label{table_domain_adaptation}
    \footnotesize
    \begin{tabular}{l|cccc}
        \toprule
        \textbf{Domain Adaptation Task} & \textbf{VICReg} & \textbf{Barlow Twins}& \textbf{SimCLR} &\textbf{CDSSL} \\
        \midrule
        MNIST $\to$ Digits Dataset & 67.7 &66.8 & 71.1 & \textbf{78.3} \\
         CIFAR-10 $\to$ STL10 & 65.4 & 65.2 & 67.8 & \textbf{70.2} \\
         CIFAR-10 $\to$ CIFAR-100 & 18.3 & 17.2 & 18.6  & \textbf{21.4}  \\
         CIFAR-100 $\to$ CIFAR-10 & 57.4 & 57.6 & 59.3 & \textbf{62.1} \\
        \bottomrule
    \end{tabular}%
\end{table*}

\section{Conclusion}


In this paper, we introduced CDSSL, a comprehensive framework that integrates both linear correlation and nonlinear dependence to enhance representation learning. By leveraging correlation and HSIC within a RKHS, CDSSL captures rich and diverse features while minimizing redundancy and information collapse. 
The results show that CDSSL outperforms state-of-the-art methods like VICReg and Barlow Twins, highlighting its capacity to generalize across datasets with varying distributions and complexities.

CDSSL unifies and extends existing self-supervised learning paradigms through its carefully designed loss terms, which address sample-wise and feature-wise dependencies across augmentations.

Experiments on diverse datasets, including ImageNet100, MNIST, CIFAR-10, CIFAR-100, and STL10, demonstrate the robustness and adaptability of CDSSL in various tasks, such as domain adaptation and nonlinear classification.




\section*{Acknowledgment}

This work was supported by the Natural Sciences and Engineering Research Council of Canada (NSERC) through the Discovery Grants
Program.

\ifCLASSOPTIONcaptionsoff
  \newpage
\fi

\bibliographystyle{IEEEtran}
\bibliography{references}

%

\newpage

\begin{IEEEbiography}[{\includegraphics[width=1in,height=1.25in,clip,keepaspectratio]{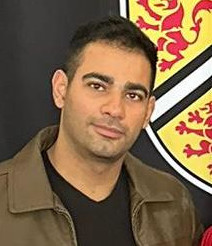}}]{M. Hadi Sepanj} is a Ph.D. candidate in the Vision and Image Processing (VIP) Lab at the University of Waterloo, Canada. He received his B.Sc. in Computer Science and M.Sc. in Artificial Intelligence before joining the Systems Design Engineering department for his doctoral studies. His research focuses on computer vision, with a particular interest in self-supervised learning, Generative Models, Statistical Machine Learning, and deep learning methods for visual understanding.
\end{IEEEbiography}

\vspace{-11cm}

\begin{IEEEbiographynophoto}{Benyamin Ghojogh} received the B.Sc. degree in electrical engineering from the Amirkabir University of Technology, Tehran, Iran, in 2015, the M.Sc. degree in electrical engineering from the Sharif University of Technology, Tehran, Iran, in 2017, and the Ph.D. in electrical and computer engineering (in the area of pattern analysis and machine intelligence) from the University of Waterloo, Waterloo, ON, Canada, in 2021. He was a postdoctoral fellow, focusing on machine learning, at the University of Waterloo, in 2021. His research interests include machine learning, dimensionality reduction, manifold learning, computer vision, data science, and deep learning.
\end{IEEEbiographynophoto}

\vspace{-11cm}


\begin{IEEEbiography}[{\includegraphics[width=1in,height=1.25in,clip,keepaspectratio]{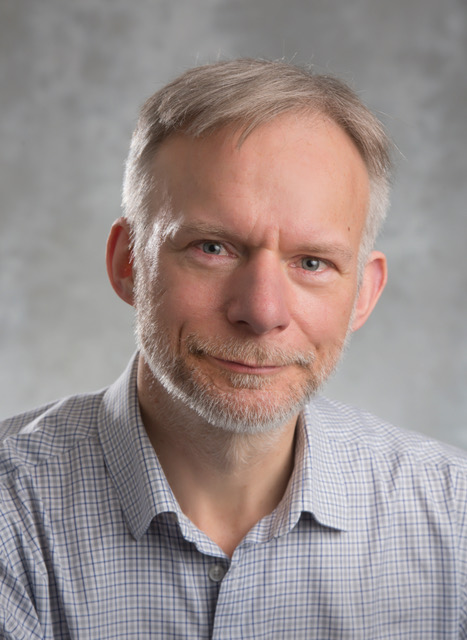}}]{Paul Fieguth} studied undergraduate Electrical Engineering at the University of Waterloo and graduate engineering degrees at the Massachusetts Institute of Technology (MIT).  He has been a member of the faculty at the University of Waterloo in Systems Design Engineering since 1996, where he has been Associate Chair Undergraduate, Department Chair, Associate Dean and, since 2023, Associate Vice President.
His research interests include statistical signal and image processing, hierarchical algorithms, data fusion, machine learning, and the interdisciplinary applications of such methods.  He has significant pedagogical interests in the area of complex systems, specifically developing a much deeper understanding among engineering students on the impact of complex systems in many areas of engineering decision making.  He is the author of three textbooks, a 2010 text on Statistical Image Processing \& Multidimensional Modeling, a 2021 text on Complex Systems, and a 2022 text on Pattern Recognition.
\end{IEEEbiography}




\end{document}